\documentclass{bmvc2k}
\usepackage{makecell}
\usepackage{siunitx}     
\usepackage{tabularx}
\usepackage{graphicx}
\usepackage{booktabs}
\usepackage{adjustbox}
\usepackage{amssymb}  
\usepackage{pifont}   
\usepackage{xcolor}
\usepackage{multirow}
\usepackage[ruled]{algorithm2e}
\usepackage{wrapfig}
\usepackage{float}
\usepackage{footmisc}
\title{CellMamba: Adaptive Mamba for Accurate and Efficient Cell Detection}

\addauthor{Ruochen Liu$^{\textstyle *}$}{sgrliu18@liverpool.ac.uk}{1}
\addauthor{Yi Tian$^{\textstyle *}$}{e1561428@u.nus.edu}{2}
\addauthor{Jiahao Wang}{Jiahao.Wang16@student.xjtlu.edu.cn}{3}
\addauthor{Hongbin Liu}{Hongbin.Liu@xjtlu.edu.cn}{3}
\addauthor{Xianxu Hou}{Xianxu.Hou@xjtlu.edu.cn}{3}
\addauthor{Jingxin Liu\textsuperscript{$\boldsymbol \dagger$}}{Jingxin.Liu@xjtlu.edu.cn}{3}

\addinstitution{
 Faculty of Science and Engineering\\
 University of Liverpool\\
 Liverpool, UK
}
\addinstitution{
 Yong Loo Lin School of Medicine\\
 National University of Singapore\\
 Singapore
}
\addinstitution{
School of AI and Advanced Computing\\
Xi'an Jiaotong-Liverpool University\\
Suzhou, China
}

\runninghead{Liu et al.}{CellMamba: Adaptive Mamba for Cell Detection}

\begin{document}
\footnotetext{$^{\textstyle *}$Equal Contribution; \textsuperscript{$\boldsymbol \dagger$}Corresponding Author}
\maketitle

\begin{abstract}
Cell detection in pathological images presents unique challenges due to densely packed objects, subtle inter-class differences, and severe background clutter. In this paper, we propose CellMamba, a lightweight and accurate one-stage detector tailored for fine-grained biomedical instance detection. Built upon a VSSD backbone, CellMamba integrates CellMamba Blocks, which couple either NC-Mamba or Multi-Head Self-Attention (MSA) with a novel Triple-Mapping Adaptive Coupling (TMAC) module. TMAC enhances spatial discriminability by splitting channels into two parallel branches, equipped with dual idiosyncratic and one consensus attention map, adaptively fused to preserve local sensitivity and global consistency. Furthermore, we design an Adaptive Mamba Head that fuses multi-scale features via learnable weights for robust detection under varying object sizes. Extensive experiments on two public datasets—CoNSeP and CytoDArk0—demonstrate that CellMamba outperforms both CNN-based, Transformer-based, and Mamba-based baselines in accuracy, while significantly reducing model size and inference latency. Our results validate CellMamba as an efficient and effective solution for high-resolution cell detection.
\end{abstract}

%-------------------------------------------------------------------------
\section{Introduction}
\label{sec:intro}
Nucleus/cell detection is a fundamental task in pathological diagnosis and microscopic image analysis \cite{Pan2018,afifi2025,soliman2024}. By analysing pathological images, subtle morphological changes in cells can be accurately identified, providing critical insights for disease diagnosis \cite{Srinidhi2021Deep}. However, interpreting whole slide images (WSIs) requires extensive expertise and is inherently time-consuming \cite{Ayyad2021,Srinidhi2021Deep}. In recent years, advancements in computational pathology (CPath) have facilitated the integration of artificial intelligence (AI) algorithms into computer-aided diagnosis (CAD) systems \cite{Hosseini2024Computational}. These AI-driven approaches not only improve diagnostic accuracy but also yield substantial reductions in the time and economic costs inherent to the entire diagnostic and therapeutic workflow \cite{VanderLaak2021Deep, Srinidhi2021Deep,cui2021}.

Deep learning-based nucleus and cell detection methods can generally be categorised into three approaches: \emph{instance segmentation}, \emph{point-based detection}, and \emph{bounding-box-based detection}. \emph{Instance segmentation} delineates individual cells with pixel-level precision, explicitly assigning semantic categories to each cellular instance while capturing detailed structural information \cite{graham2019hover,horst2024cellvit, vadori2024cisca, TAJBAKHSH2020}. However, it requires labour-intensive annotations and is computationally demanding, making it less practical for large-scale pathological analysis \cite{TAJBAKHSH2020,Huang2023}. \emph{Point-based detection} represents each nucleus with a single point that denotes the centroid and is explicitly linked to the category, significantly reducing computational complexity and annotation effort \cite{Huang2023,DPA-P2PNet,shui_cell, Shui2025Towards}. While it effectively achieves classification and rough location capture, it lacks information on cell size and boundaries \cite{Huang2023}, limiting its applicability to tasks requiring detailed morphological analysis, which is critical for the diagnosis and prognostic evaluation of various diseases \cite{Srinidhi2021Deep}. \emph{Bounding-box-based detection} delivers complete, rectangular approximated cellular regions and cell-level information, with lower annotation requirements \cite{zhu2024}. It can both reflect cell position and morphological information without requiring heavy annotation, as its bounding-boxes can be automatically generated based on point annotations or pixel-level annotations. However, research on this approach remains limited, as most bounding-box-based cell detection tasks still rely on general object detection models rather than pathology-specific solutions. 

Compared to natural images, pathology images present unique challenges: target cells are small, densely distributed, exhibit varied shapes and staining patterns, and generally have low contrast against complex tissue backgrounds, collectively raising the difficulty of cell detection \cite{stringer2021cellpose,soliman2024, graham2019hover}. Meanwhile, subtle inter-class differences further complicate cell classification by hindering accurate discrimination. CNN-based models like HoVer-Net \cite{graham2019hover} capture fine-grained local textures to extract salient cues from minute details but are limited by narrow receptive fields, while Transformer-based approaches such as CellViT \cite{horst2024cellvit} offer global context at the cost of high computation, particularly for dense predictions \cite{mcgenity2024}. Vision Mamba models have recently shown strong potential for efficient visual representation, combining long-range dependency modelling with linear complexity \cite{gu2023mamba, Liu2024ViM, vmamba2024}. Mamba-2 \cite{dao2024transformers}, in particular, leverages hardware-friendly operations to outperform traditional attention mechanisms in efficiency. Despite this, Mamba-based architecture remains underexplored in pathology, where its lightweight design could be well-suited for dense, fine-grained tasks like cell detection.

To tackle the challenges of efficient and accurate cell detection, we propose CellMamba, a one-stage detection framework based on Mamba for pathology images. CellMamba features a mixed Mamba-Transformer backbone and an Adaptive Mamba Head. The backbone is built upon VSSD \cite{shi2024vssd}, which leverages Non-Causal State-Space Duality (NC-SSD) extended from Mamba-2 for hierarchical visual representation, mitigating the limitations of causal modelling. To align with the Mamba-2 terminology, we refer to this module as NC-Mamba in this paper. To further enhance detection performance without introducing substantial computational overhead, we introduce a Triple-Mapping Adaptive Coupling (TMAC) module. TMAC splits features into two parallel branches, each processed with either NC-Mamba or Multi-Head Self-Attention (MSA) \cite{vaswani2017attention}, enabling specialisation in complementary visual cues such as texture and boundaries. The outputs are fused via adaptive attention mapping coupling, improving focus on nuclei and suppressing irrelevant background noise. Additionally, we design an Adaptive Mamba Head for multi-scale detection, to better capture cells of varying sizes while maintaining low latency. The main contributions of this paper are summarised as follows:
\begin{enumerate}
    \item We propose CellMamba, a Mamba-based one-stage object detector for efficient and accurate cell detection in pathology images.
    \item We introduce the Triple-Mapping Adaptive Coupling (TMAC) module to reduce feature interference and enhance spatial focus, improving detection in dense, low-contrast regions.
    \item We bridge the gap in bounding-box-based cell detection with Mamba models and validate our method on two public datasets, achieving state-of-the-art performance.
\end{enumerate}

\section{Related Work}
\textbf{CNN-Based Methods.} CNNs have long been the backbone of pathological image analysis. U-Net \cite{Ronneberger2015UNet} introduced an encoder–decoder design with skip connections for efficient segmentation, establishing a foundational architecture that subsequent segmentation models have widely adopted. RetinaNet \cite{lin2017focal} and Mask R-CNN \cite{he2017mask} extended CNNs to detection and instance segmentation, with strong performance on class-imbalanced samples and small objects \cite{da2022digestpath}. HoVer-Net \cite{graham2019hover} improved nuclear instance separation via additional horizontal and vertical distance maps, while DoNet \cite{Jiang2023DoNet} addressed overlapping cytoplasm using mask-guided decomposition. However, CNNs are limited in modelling global context, which is critical in complex pathology images.

\textbf{Transformer-Based Methods.} Transformers leverage self-attention \cite{vaswani2017attention} for long-range dependency modelling and have shown promise in medical imaging. ViT \cite{Dosovitskiy2021ViT} pioneered a pure Transformer for vision via patch-based tokenisation but demands large-scale data and computation. DETR \cite{carion2020end} introduced Transformers to end-to-end detection, with Deformable DETR \cite{zhu2020deformable} and DINO \cite{zhang2022dino} further improving detection efficiency and performance. Specifically in pathology, CellViT \cite{horst2024cellvit} demonstrated strong generalisation, but Transformers remain computationally expensive, limiting their practicality in dense and high-resolution settings.

\textbf{Mamba-Based Methods.} Mamba \cite{gu2023mamba}, based on State Space Models (SSMs) \cite{gu2021efficiently, gu2021combining, smith2022simplified}, offers efficient long-range modeling with linear complexity. Its successor, Mamba-2 \cite{dao2024transformers} introduces State Space Duality (SSD) for better hardware utilisation. Vision extensions include ViM \cite{Liu2024ViM} and VMamba \cite{vmamba2024}, which adopt multi-directional scanning for 2D image modelling. VSSD \cite{shi2024vssd} further introduces a non-causal formulation to remove the constraints of sequential scanning, enabling more flexible and efficient vision representation. Detection models like Mamba-YOLO \cite{wang2024mamba}, MobileMamba \cite{he2024mobilemamba}, and Spatial-Mamba \cite{xiao2024spatial} enhance efficiency and spatial reasoning. In medical imaging, Mamba variants have been used for segmentation and classification \cite{zhang20242dmamba, mambaunet2024, ultralight2024, s3mamba2024}, but their use in instance-level cell detection remains largely unexplored.

\section{Method}

\begin{figure}[t]
\includegraphics[width=\textwidth]{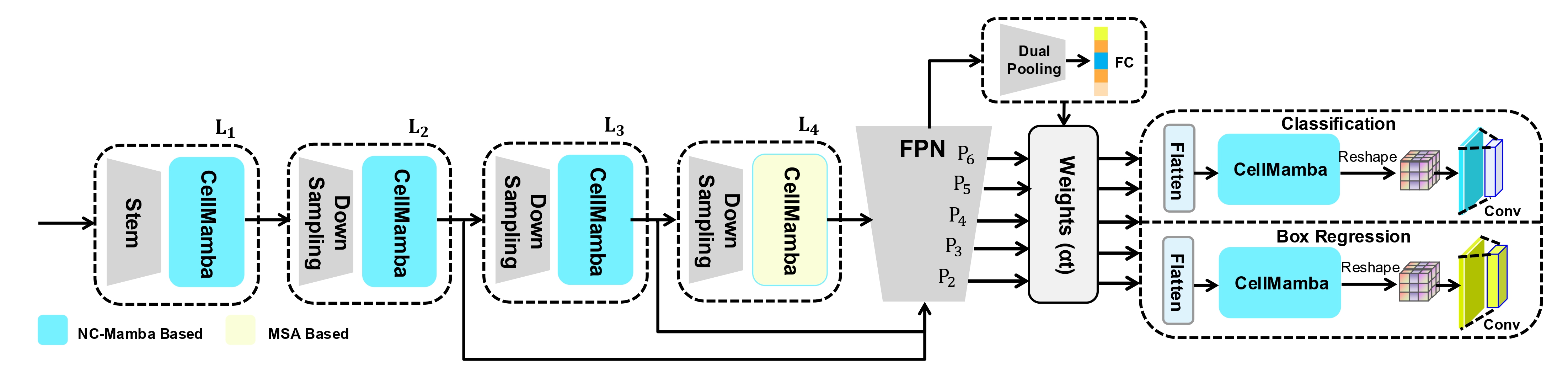}
\caption{The overall framework architecture of CellMamba, comprises a four-stage mixed Mamba-Transformer hierarchical backbone and an adaptive Mamba head for cell classification and box regression.} 
\label{fig1}
\end{figure}

We present CellMamba, a one-stage object detection framework fully built on Mamba for efficient and accurate cell detection. It combines a hierarchical Mamba backbone incorporating triple-mapping adaptive coupling modules, and an adaptive Mamba head. The overall architecture is shown in Figure ~\ref{fig1}.

\subsection{Backbone Architecture}

Our backbone is based on the VSSD framework~\cite{shi2024vssd}, which we adapt for cell detection through task-specific modifications that improve fine-grained localization and efficiency. These adjustments address the challenges posed by morphological similarity and dense cell distributions in pathology images.

The first three stages of the backbone use NC-Mamba blocks to capture long-range spatial dependencies, which are essential for separating visually similar nuclei and suppressing background noise. The final stage adopts Multi-Head Self-Attention (MSA)~\cite{vaswani2017attention} to enhance global contextual modelling and complex spatial reasoning for low-resolution abstract cellular features. 

We refer to each stage’s combined structure of sequence modeling (NC-Mamba or MSA) followed by Triple-Mapping Adaptive Coupling (TMAC) as a CellMamba block, which serves as the core unit of our backbone. The four stages contain 2, 2, 8, and 4 such blocks respectively, each equipped with our proposed TMAC module, which will be detailed in the following subsection.

To strengthen multi-scale feature interaction and improve detection across varying cell sizes and densities, we incorporate a Feature Pyramid Network (FPN)~\cite{lin2017feature}. Outputs from stages $\text{L}_2$–$\text{L}_4$ are fused into five feature maps $\text{P}_2$-$\text{P}_6$, enabling the propagation of both semantic and spatial information for precise, scale-aware cell detection.

\subsection{Triple-Mapping Adaptive Coupling}

\begin{figure}[t]
\centering
\includegraphics[width=0.9\linewidth]{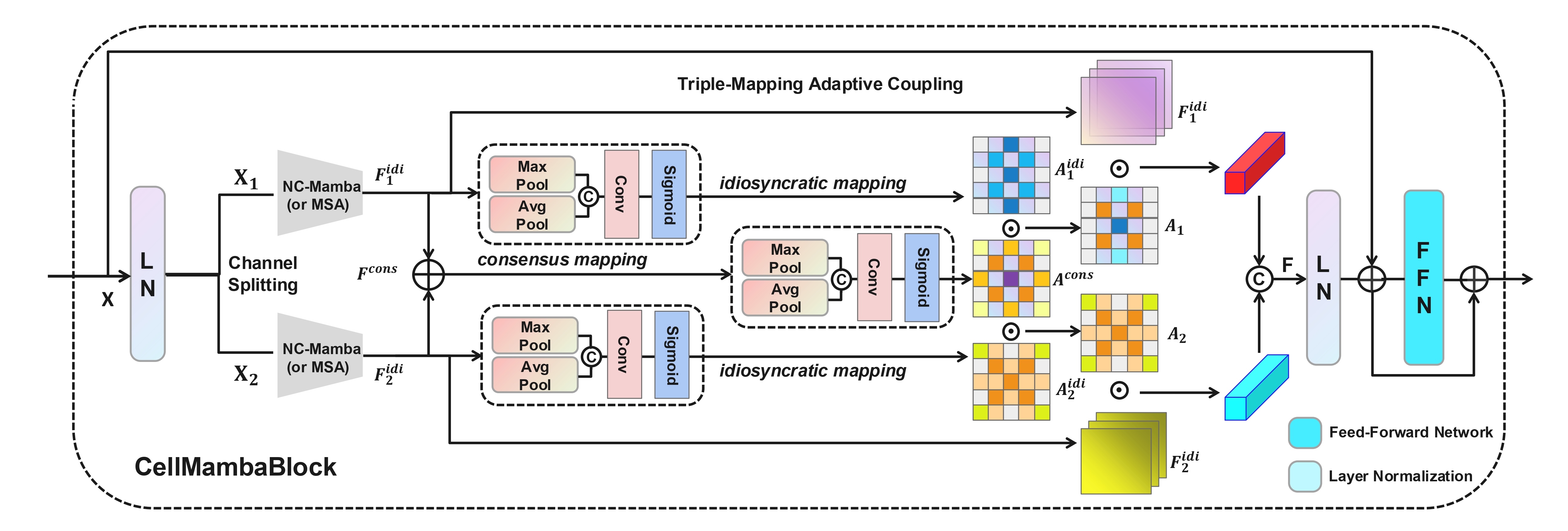}
\caption{Detailed structure of the proposed CellMamba block. TMAC is placed after sequence attention and before Feed-Forward Network (FFN), enabling local spatial refinement via triple mapping and adaptive coupling.}
\label{fig:cellmamba_block}
\end{figure}

While sequence-based models like NC-Mamba and MSA are effective at modeling long-range dependencies, they often overlook localized spatial cues, such as nuclear boundaries and texture variations,that are critical for accurate cell detection. To address this limitation, we introduce the Triple-Mapping Adaptive Coupling (TMAC) module, which augments each attention block with a lightweight yet expressive spatial refinement mechanism. By incorporating dual-path channel splitting and triple mapping, TMAC enables fine-grained spatial focus and robust feature alignment. As illustrated in Figure~\ref{fig:cellmamba_block}, it is inserted between the sequence attention layer and Feed-Forward Network (FFN) of each CellMamba block.

\paragraph{Channel Splitting}  
To enhance spatial specialisation and reduce interference among heterogeneous features, we adopt a Dual-Channel structure by splitting the input feature map along the channel dimension. Given an input $\mathbf{X} \in \mathbb{R}^{H \times W \times C}$, we flatten and split it into two sub-paths:
\begin{equation}
\{\mathbf{X}_1, \mathbf{X}_2\} = \text{Split}(flatten(\mathbf{X})), \quad \mathbf{X}_m \in \mathbb{R}^{B \times L \times \frac{C}{2}}, \quad m \in \{1,2\}
\end{equation}

Each sub-path can focus on complementary morphological cues—such as intra-cell textures or inter-cell boundaries—while sharing the same sequence modelling backbone (NC-Mamba or MSA). This structure improves feature disentanglement and reduces computation and memory \cite{ultralight2024}, which is critical for high-resolution pathology inference. The detailed spatial refinement of each sub-path is then handled by our proposed TMAC module.

\paragraph{Triple Mapping}
TMAC generates two idiosyncratic attention maps $\mathbf{A}_m^{idi}$ (one for each sub-path) and one consensus attention map $\mathbf{A}^{cons}$ shared across branches. Given the sub-path features $\mathbf{F}_m \in \mathbb{R}^{B \times L \times \frac{C}{2}}$, we reshape them into 2D form $\mathbf{F}_m^{idi} \in \mathbb{R}^{H \times W \times \frac{C}{2}}$ and compute:

\begin{equation}
\mathbf{A}_m^{idi} = \sigma \left(\text{Conv}\left(\text{concat}\left[
\underset{\frac{C}{2}}{\text{mean}}(\mathbf{F}_m^{idi}), \underset{\frac{C}{2}}{\text{max}}(\mathbf{F}_m^{idi})
\right] \right) \right), \quad m \in \{1,2\}
\end{equation}

To capture common regions of interest, we compute the consensus feature map by summing the sub-path features:
\begin{equation}
\mathbf{F}^{cons} = \sum_{m=1}^{2} \mathbf{F}_m^{idi}, \quad \mathbf{F}^{cons} \in \mathbb{R}^{H \times W \times \frac{C}{2}}
\end{equation}

The consensus attention map is then computed using the same pooling and convolutional process:
\begin{equation}
\mathbf{A}^{cons} = \sigma \left(\text{Conv}\left(\text{concat}\left[\underset{\frac{C}{2}}{\text{mean}}(\mathbf{F}^{cons}), \underset{\frac{C}{2}}{\text{max}}(\mathbf{F}^{cons}) \right] \right) \right)
\end{equation}

All three attention branches share weights to ensure consistency and reduce complexity. While sharing parameters, the attention maps retain distinct focuses due to different input distributions, achieving a balance between diversity and efficiency.

\paragraph{Adaptive Coupling}
To avoid early-stage interference from immature features, we adopt a dynamic coupling strategy, which is divided into two stages based on training progress. During the first $N$ epochs, when the loss is still dropping sharply, we disable consensus fusion by setting $\mathbf{A}^{cons}$ to an all-ones matrix, allowing each sub-path to independently learn spatial saliency without interference. This stage encourages the specialization of $\mathbf{A}_m^{idi}$ toward path-specific cues such as texture or boundary structures.

When the loss levels off and the idiosyncratic attention maps stabilize on the dataset from epoch $N$ onward, we activate full coupling by fusing the idiosyncratic $\mathbf{A}_m^{idi}$ and consensus attention maps $\mathbf{A}^{cons}$ through gated element-wise multiplication:

\begin{equation}
\left\{ \mathbf{A}_m \right\}_{ m \in \{1,2\}} = \left\{ \mathbf{A}_m^{idi} \odot \mathbf{A}^{cons} \right\}_{m \in \{1,2\}},
\quad \mathbf{A}_m,\mathbf{A}_m^{idi}, \mathbf{A}^{cons} \in \mathbb{R}^{H \times W \times 1}
\end{equation}

This formulation ensures that consensus guidance is introduced only after local specialization is reliable, promoting convergence stability and enabling the network to emphasize both fine-grained branch-specific details and globally consistent regions.

\paragraph{Feature Fusion}
Each attention map is broadcast across channels and used to modulate its corresponding feature map:
\begin{equation}
\mathbf{F}_m^{\text{final}} = Broadcast(\mathbf{A}_m, channel=\frac{C}{2}) \odot \mathbf{F}_m^{\text{idi}}, \quad m \in \{1, 2\}
\end{equation}
The final feature representation $\mathbf{F}$ is obtained by flattening and concatenating the two branches, and is subsequently fed into the Layer Normalization (LN) and Feed-Forward Network (FFN):
\begin{equation}
\mathbf{F} = \text{concat}\left(flatten( \mathbf{F}_1^{\text{final}}), flatten(\mathbf{F}_2^{\text{final}}) \right)
\end{equation}

TMAC enhances discriminative power by refining spatial attention in a branch-specific and dynamically coupled manner. This design directly supports cell detection tasks, where accurate localization of small, overlapping, and morphologically similar nuclei is crucial. By aligning local attention with global semantic cues, TMAC improves focus on fine-grained structures while suppressing noise in dense pathological imagery.

\subsection{Adaptive Mamba Head} 
Due to Mamba's continuous state updates, it possesses stronger global modeling capability compared to convolutions. Leveraging this, we design a lightweight detection head by employing a single \textbf{CellMamba block} in both the classification and box regression branches. 

In the Detection Head, we integrate multiple classification and box regression branches to detect cells across different scales, with final target aggregation. Recognizing lower-level FPN features capture fine cellular structures/details while higher-level features localize dense regions, we design a dynamic weight mechanism for adaptive multi-scale detection.

The FPN outputs $\{\text{P}_i \}_{i=2}^6$ aggregate to $x \in \mathbb{R}^{T \times H \times W \times C}$ with T denoting the number of output feature maps (T=5), which then undergoes dual-pooling:  

\begin{enumerate}
    \item For each element $x_t$ in the T dimension ($t \in [1,5]$) and each channel $c$ ($c \in [1,C]$), global pooling is first performed in the spatial dimension ($H\!\times\!W$) : 
    \[
    \displaystyle z_t^c = \frac{1}{H \times W} \sum_{i=1}^H \sum_{j=1}^W x_t^c(i, j), \quad z_t^c \in \mathbb{R}^{T \times C}
    \]

    \item Secondary pooling is performed in the channel dimension ($C$):  
    \[
    \displaystyle s_t = \frac{1}{C} \sum_{c=1}^C z_t^c, \quad s_t \in \mathbb{R}^{T \times 1}
    \]
\end{enumerate}

Inter-dependencies across T are modeled via fully connect layer (FC),  where $\alpha_t \in \mathbb{R}^{T \times 1}$ denotes the weight for the $t$-th feature map:
\[\alpha_t = \text{Sigmoid}\big(\text{FC}(s_t)\big)\] Finally, the classification and box regression branches operate on the weighted feature maps $\alpha_t \text{P}_i$ ($i \in [2,6]$) for varying cell sizes and densities, defined as:
\begin{equation}
\text{Class\_sub} = \text{Conv}_1(\text{Reshape}(\text{CellMamba}(flatten(\alpha_t\text{P}_i)))) \in \mathbb{R}^{K \times H' \times W'},\quad i \in [2,6]
\end{equation}
\begin{equation}
\text{Box\_sub} = \text{Conv}_2(\text{Reshape}(\text{CellMamba}(flatten(\alpha_t\text{P}_i)))) \in \mathbb{R}^{4 \times H' \times W'},\quad i \in [2,6]
\end{equation}

By enriching scale-specific features with long-range semantics, the head improves cell classification and boundary localization while maintaining high efficiency.

\textbf{Loss Function}
We adopt Focal Loss~\cite{lin2017focal} for classification to mitigate class imbalance, and Smooth L1 Loss~\cite{girshick2015fast} for bounding box regression, with their weighted sum used as the total objective.

\section{Experiments}
\label{sec:experiments}

\subsection{Datasets}

We evaluate our method on two publicly available histopathological datasets, CoNSeP \cite{graham2019hover} and CytoDArk0 dataset~\cite{vadori2024cisca}, adapting them for bounding-box-based cell detection. The CoNSeP dataset consists of H\&E-stained colorectal adenocarcinoma images with instance segmentation annotations, which we convert into bounding boxes and categorise into four classes for detection and classification evaluation. To balance efficiency and resolution, we extract 128$\times$128 patches from whole-slide images. The CytoDArk0 comprises Nissl-stained brain tissue images at 40$\times$ magnification. We generate bounding boxes from segmentation masks, framing it as a single-cell detection task, and extract 256$\times$256 patches to assess performance on brain tissue images. While CoNSeP is designed for nucleus detection and CytoDArk0 for cell detection, evaluating the model on these two datasets fully validates its capabilities across multiple pathological detection tasks.

\subsection{Implementation Details}
We assess detection performance using three widely adopted metrics: mean Average Precision (mAP), mAP@50, and mAP@75. Additionally, we perform post-processing on the detection results, determine the optimal confidence threshold, and compute the corresponding Macro-averaged Precision, Recall, and F1-score to ensure a fair comparison with semantic segmentation models. To assess computational efficiency, we measure Inference Time and Number of Parameters. For training, we adopt the SGD optimiser with an initial learning rate of \(1\times10^{-3}\) and weight decay of \(1\times10^{-4}\). The learning rate is adjusted using a combination of LinearLR and MultiStepLR. Through experimental validation and quantitative observation, it is determined that setting \( N = 35 \) is optimal for both the CoNSeP and CytoDArk0 datasets. At this epoch, the loss functions of the model on both datasets converge to a stable plateau, indicating that the model has entered a phase of stable training without further significant loss reduction.

\subsection{Experimental Results}

\begin{table}[th]
\centering
\caption{\footnotesize Performance on CoNSeP and CytoDArk0. Bold = best, underline = second-best. M = Mask R-CNN, R = RetinaNet.}
\label{tab:results}
\resizebox{0.9\columnwidth}{!}{%
\begin{tabular}{llcccccc}
\toprule
\multirow{2}{*}{Category} &
  \multirow{2}{*}{Model} &
  \multicolumn{3}{c}{CoNSeP (mAP \%)} &
  \multicolumn{3}{c}{CytoDArk0 (mAP \%)} \\ \cline{3-8} 
                        &                                                    & mAP            & @50            & @75  & mAP  & @50  & @75  \\ \hline
\multirow{2}{*}{\emph{CNN}}    & RetinaNet~\cite{lin2017focal} (ICCV'17)            & 19.1           & 39.3           & 17.1 & 49.1 & 77.9 & 56.1 \\
                        & Mask R-CNN~\cite{he2017mask} (ICCV'17)             & 17.1           & 35.9           & 16.3 & 48.8 & 78.5 & 55.0 \\ \hline
\multirow{2}{*}{\emph{Transformer}} &
  Deformable-DETR~\cite{zhu2020deformable} (ICLR'21) &
  23.8 &
  44.2 &
  23.5 &
  47.9 &
  79.3 &
  51.7 \\
 &
  DINO~\cite{zhang2022dino} (ICLR'23) &
  24.2 &
  45.1 &
  \textbf{24.2} &
  \underline{53.0} &
  \underline{81.5} &
  \textbf{61.2} \\ \hline
\multirow{10}{*}{\emph{Mamba}} & VSSD-Micro-R~\cite{shi2024vssd} (ICCV'25)             & 24.5           & 48.2           & 23.3 & 50.5 & 80.8 & 55.5 \\
                        & VSSD-Micro-M~\cite{shi2024vssd} (ICCV'25)             & 23.1           & 44.8           & 22.5 & 47.6 & 78.3 & 50.9 \\
                        & Mamba-YOLO-Base~\cite{wang2024mamba} (AAAI'25)        & \underline{25.2} & \underline{50.7} & 23.6 & 52.3 & 81.2 & 56.6 \\
                        & MobileMamba-B1-R~\cite{he2024mobilemamba} (CVPR'25) & 22.8           & 43.9           & 22.6 & 41.7 & 72.3 & 43.9 \\
                        & MobileMamba-B1-M~\cite{he2024mobilemamba} (CVPR'25) & 23.2           & 45.3           & 22.9 & 44.1 & 74.9 & 48.2 \\
                        & 2D-Mamba-R~\cite{zhang20242dmamba} (CVPR'25)       & 16.8           & 34.2           & 15.5 & 40.1 & 68.3 & 41.5 \\
                        & 2D-Mamba-M~\cite{zhang20242dmamba} (CVPR'25)       & 20.4           & 40.2           & 19.7 & 42.8 & 74.8 & 45.6 \\
                        & Spatial-Mamba-Tiny-R~\cite{xiao2024spatial} (ICLR'25) & 21.1           & 40.8           & 20.8 & 47.7 & 77.8 & 51.2 \\
                        & Spatial-Mamba-Tiny-M~\cite{xiao2024spatial} (ICLR'25) & 22.2           & 42.5           & 20.7 & 48.3 & 78.4 & 53.9 \\
 &
  \textbf{Ours} &
  \textbf{25.7} &
  \textbf{51.1} &
  \underline{23.8} &
  \textbf{53.3} &
  \textbf{83.5} &
  \underline{59.8} \\ \bottomrule
\end{tabular}%
}
\end{table}

\paragraph{Comparison with State-of-the-art.}
We compare our proposed model with state-of-the-art (SOTA) methods across three architecture categories: CNN, Transformer, and Mamba. For CNN-based methods, RetinaNet (one-stage) and Mask R-CNN (two-stage) are early classic detection networks, widely adopted as foundational baselines in pathological detection \cite{da2022digestpath}. For transformer-based models, despite the proliferation of DETR variants, most are tailored for general object detection. We thus selected Deformable DETR and DINO for comparison. These two models are seminal in the DETR lineage, serve as foundations for most later variants, and have undergone prior validation in pathological image analysis tasks \cite{DPA-P2PNet,Pang2025CelloType}. For Mamba-based models, we selected diverse variants to validate performance across technical pathways and scenarios: VSSD, the first to introduce Mamba-2 to vision, offers foundational reference; Mamba-YOLO and MobileMamba focus on lightweight scenarios, verifying efficiency-accuracy trade-offs; 2D-Mamba explores pathological image analysis; Spatial-Mamba breaks basic Mamba's spatial limitations via explicit state-space neighborhood connectivity. Together, they cover the latest architectures, lightweight designs, and applications in pathology and complex spatial analysis. As shown in Table~\ref{tab:results}, CellMamba achieves top performance on both CoNSeP and CytoDArk0 datasets, outperforming all CNN and Mamba baselines, and matching or surpassing the best Transformer-based models.

On CoNSeP, which features densely packed, morphologically similar nuclei across multiple classes, our model achieves the highest mAP (25.7\%) and mAP@50 (51.1\%), and ranks second in mAP@75 (23.8\%), just 0.4\% below DINO. These results indicate strong localization and discriminative capabilities under fine-grained and crowded conditions. On CytoDArk0, a single-class detection task with larger and less ambiguous targets, CellMamba again leads with the highest mAP (53.3\%) and mAP@50 (83.5\%), demonstrating strong generalization across different spatial contexts. We observe that certain Mamba-based models typically underperform in pathological image detection, primarily due to the far greater complexity of pathological images compared to natural images. Despite their robust feature extraction capabilities on natural images, these models lack adaptability to the unique challenges of pathological scenarios. Even 2D-Mamba, though tailored for pathological images, is originally designed for gigapixel Whole Slide Image (WSI) classification tasks; it inherently lacks high-precision local detail capture capabilities, thus still exhibiting suboptimal performance in detection tasks.

\begin{wraptable}{R}{0.6\textwidth} 
\centering
\caption{\footnotesize Precision (P), Recall (R), and F1-score on CoNSeP and CytoDArk0. Bold = best, underline = second-best.}
\vspace{2mm}
\label{tab:prf1}
\resizebox{0.6\textwidth}{!}{
\setlength{\tabcolsep}{4pt}  
\scriptsize  
\begin{tabular}{llccc|ccc}
\toprule
Category & Model & \multicolumn{3}{c|}{CoNSeP} & \multicolumn{3}{c}{CytoDArk0} \\
& & P & R & F1 & P & R & F1 \\
\midrule
\emph{CNN} & HoVer-Net~\cite{graham2019hover} & 57.2 & 55.8 & 56.5 & 82.0 & 81.8 & 81.9 \\
\emph{Transformer} & CellViT~\cite{horst2024cellvit} & \textbf{63.2} & \underline{59.9} & \underline{61.5} & \underline{83.4} & \textbf{83.0} & \underline{83.2} \\
\emph{Mamba} & \textbf{Ours} & \underline{62.2} & \textbf{67.4} & \textbf{64.8} & \textbf{86.9} & \underline{81.9} & \textbf{84.4} \\
\bottomrule
\end{tabular}
}
\end{wraptable}

We also evaluate the detection performance against HoVer-Net and CellViT, two representative instance segmentation-based cell detectors. CellMamba achieves the highest F1-scores on both datasets (Table~\ref{tab:prf1}), confirming its instance-level discriminative strength even under segmentation-style evaluation.

\setlength\intextsep{0pt}
\begin{wraptable}{L}{0.58\textwidth}
\centering
\caption{\footnotesize Model size, inference time, and detection performance (mAP@50) on CytoDArk0 (per 256$\times$256 patch). Bold = best, underline = second-best. M = Mask R-CNN, R = RetinaNet. $*$ denotes that the official version incorporates additional Test-Time Augmentation (TTA) for RetinaNet, resulting in relatively longer inference time.}
\vspace{2mm}
\label{tab:params_time}
\resizebox{0.6\textwidth}{!}{
\setlength{\tabcolsep}{4pt} 
\begin{adjustbox}{max width=0.68\textwidth}
\begin{tabular}{lccc}
\toprule
Model & Params (M) & Time (ms) & mAP@50 \\
\hline
\multicolumn{4}{l}{\textit{CNN}} \\
\hline
RetinaNet~\cite{lin2017focal} (ICCV'17)       & 36.0 & 3.4 & 77.9 \\
Mask R-CNN~\cite{he2017mask} (ICCV'17)        & 44.0 & 4.3 & 78.5 \\
\hline
\multicolumn{4}{l}{\textit{Transformer}} \\
\hline
Deformable-DETR~\cite{zhu2020deformable} (ICLR'21) & 40.0 & 1.8 & 79.3 \\
DINO~\cite{zhang2022dino} (ICLR'23)          & 48.0 & 4.5 & 81.5 \\
\hline
\multicolumn{4}{l}{\textit{Mamba}} \\
\hline
VSSD-Micro-R~\cite{shi2024vssd} (ICCV'25)         & 21.4 & 2.1 & 80.8 \\
VSSD-Micro-M~\cite{shi2024vssd} (ICCV'25)         & 33.0 & 5.3 & 78.3 \\
Mamba YOLO-Base~\cite{wang2024mamba} (AAAI'25)    & 22.0 & 2.2 & 81.2 \\
MobileMamba-B1-R~\cite{he2024mobilemamba} (CVPR'25) & 27.1 & $6.6^{*}$ & 72.3 \\
MobileMamba-B1-M~\cite{he2024mobilemamba} (CVPR'25) & 38.0 & 6.5 & 74.9 \\
2D-Mamba-R~\cite{zhang20242dmamba} (CVPR'25)   & 38.8 & 2.4 & 68.3 \\
2D-Mamba-M~\cite{zhang20242dmamba} (CVPR'25)   & 49.0 & 6.9 & 74.8 \\
Spatial-Mamba-Tiny-R~\cite{xiao2024spatial} (ICLR'25) & 36.3 & 5.8 & 77.8 \\
Spatial-Mamba-Tiny-M~\cite{xiao2024spatial} (ICLR'25) & 46.0 & 6.2 & 78.4 \\
\textbf{Ours}                                  & \textbf{14.7} & \textbf{1.6} & \textbf{83.5} \\
\bottomrule
\end{tabular}
\end{adjustbox}
}
\end{wraptable}

\paragraph{Efficiency Analysis.}
Table~\ref{tab:params_time} summarises inference time, model size, and performance. Our model achieves the best overall trade-off—only 14.7M parameters and 1.6 ms latency per 256×256 patch,while maintaining the highest mAP@50. Compared to models like CellViT (approximately 3.4× larger) and DINO (approximately 3.3× larger, 2.8× slower), CellMamba is significantly more efficient. Some Mamba-based baselines demonstrate certain competitiveness and potential, but still lack our combination of speed, compactness, and accuracy.

Overall, our method balances performance and efficiency across diverse detection settings: excelling in multi-class fine-grained discrimination, scaling well to large object datasets, and maintaining minimal inference cost.

\paragraph{Ablation Study.}
As shown in Table~\ref{tab:ablation_study}, we conduct ablation experiments on the CoNSeP dataset to evaluate the individual contributions of CellMamba’s key components. To ensure intuitive comparison with the previous Table~\ref{tab:params_time}, we uniformly resized CoNSeP to a resolution of 256×256 when calculating inference time and parameters. Starting from the VSSD backbone, we first apply the channel splitting strategy, which slightly improves mAP (+0.1\%) while reducing both inference time and parameter count, confirming its effectiveness in lightweight and disentangled representation learning. Incorporating the Triple-Mapping Adaptive Coupling (TMAC) module yields a more substantial gain (+0.7\% mAP and +1.9\% mAP@50) without introducing overhead, demonstrating its strength in refining spatial attention and emphasizing subtle nuclear structures. Finally, replacing the original detection head with an adaptive Mamba head further improves accuracy and reduces complexity, validating the benefit of lightweight sequence modeling in high-resolution pathology.

These results confirm the rationale behind our modular design: each component addresses a distinct challenge—channel interference, spatial alignment, or contextual modeling—contributing to a compact yet highly discriminative detection framework tailored for dense and morphologically complex cell imagery.

\begin{table}[t]
\centering
\caption{\footnotesize Ablation study based on the VSSD backbone on the CoNSeP dataset. TMAC = Triple-Mapping Adaptive Coupling. Bold = best.}
\label{tab:ablation_study}
\resizebox{1\columnwidth}{!}{
\begin{tabular}{lccccc}
\toprule
\textbf{Model Variant} & mAP & mAP@50 & mAP@75 & Time (ms) & Params (M) \\
\midrule
VSSD Backbone & 24.5 & 48.2 & 23.3 & 2.1 & 21.4 \\
VSSD Backbone + Channel Splitting & 24.6 & 48.5 & 23.3 & 1.7 & 18.1 \\
VSSD Backbone + Channel Splitting + TMAC & 25.3 & 50.4 & 23.7 & 1.7 & 18.1 \\
VSSD Backbone + Channel Splitting + TMAC + Adaptive Mamba Head & \textbf{25.7} & \textbf{51.1} & \textbf{23.8} & \textbf{1.6} & \textbf{14.7} \\
\bottomrule
\end{tabular}
}
\end{table}

\section{Conclusion}

In this paper, we propose CellMamba, a lightweight and accurate one-stage cell detector designed for high-resolution pathological images. The architecture combines a hierarchical backbone with the proposed Triple-Mapping Adaptive Coupling (TMAC) module, which enhances spatial focus through complementary attention maps and adaptive fusion. In addition, an adaptive Mamba head improves detection across varied cell sizes and densities. Experimental results on CoNSeP and CytoDArk0 demonstrate that CellMamba achieves state-of-the-art accuracy while maintaining fast inference and compact model size. These findings underscore the effectiveness of structured state-space models in dense and fine-grained biomedical detection and offer a scalable framework for future research in computational pathology.\\

\noindent \textbf{Acknowledgments} \ This work was jointly supported by the National Natural Science Foundation of China (62201474 and 62206180), Suzhou Science and Technology Development Planning Programme (Grant No.ZXL2023171) and XJTLU Research Development Fund (RDF-21-02-084, RDF-22-01-129 and RDF-23-01-053).

\bibliography{bmvc_final}
\end{document}